\documentclass[11pt,a4paper]{article}

\usepackage[utf8]{inputenc}
\usepackage[T1]{fontenc}
\usepackage{lmodern}
\usepackage{microtype}

\usepackage{amsmath,amssymb,amsfonts}
\usepackage{mathtools}

\usepackage{booktabs}
\usepackage{array}
\usepackage{multirow}
\usepackage{graphicx}

\usepackage[margin=1in]{geometry}
\usepackage{setspace}
\usepackage{parskip}

\usepackage{listings}
\usepackage[colorlinks=true,linkcolor=blue,citecolor=blue,urlcolor=blue]{hyperref}
\usepackage{xurl} 

\usepackage{natbib}
\title{Semantic Compression of LLM Instructions via Symbolic Metalanguages}
\author{Ernst van Gassen\\
  \textit{Independent Researcher}\\
  \texttt{evg.gassen@gmail.com}
}
\date{}

\begin{document}

\maketitle

\begin{abstract}
We introduce MetaGlyph, a symbolic language for compressing prompts by encoding instructions as mathematical symbols rather than prose. Unlike systems requiring explicit decoding rules, MetaGlyph uses symbols like $\in$ (membership) and $\Rightarrow$ (implication) that models already understand from their training data. We test whether these symbols work as ``instruction shortcuts'' that models can interpret without additional teaching.

We evaluate eight models across two dimensions relevant to practitioners: scale (3B--1T parameters) and accessibility (open-source for local deployment vs.\ proprietary APIs). MetaGlyph achieves 62--81\% token reduction across all task types. For API-based deployments, this translates directly to cost savings; for local deployments, it reduces latency and memory pressure.

Results vary by model. Gemini 2.5 Flash achieves 75\% semantic equivalence between symbolic and prose instructions on selection tasks, with 49.9\% membership operator fidelity. Kimi K2 reaches 98.1\% fidelity for implication ($\Rightarrow$) and achieves perfect (100\%) accuracy on selection tasks with symbolic prompts. GPT-5.2 Chat shows the highest membership fidelity observed (\textbf{91.3\%}), though with variable parse success across task types. Claude Haiku 4.5 achieves 100\% parse success with 26\% membership fidelity. Among mid-sized models, Qwen 2.5 7B shows 62\% equivalence on extraction tasks. Mid-sized open-source models (7B--12B) show near-zero operator fidelity, suggesting a U-shaped relationship where sufficient scale overcomes instruction-tuning biases.
\end{abstract}


\section{Introduction}

Large language models are controlled through natural-language prompts. These prose instructions specify tasks and constraints, and the approach works well enough to be widely adopted. But it has two weaknesses. First, prompts are verbose: expressing precise constraints takes many words. Second, prompt behavior is fragile: small wording changes can produce very different outputs.

These problems have real costs. More tokens mean higher inference costs and latency. Ambiguous phrasing leads to inconsistent results. Yet models are trained on vast amounts of mathematical and logical notation, where compact symbols carry precise meaning. This raises a natural question: can we communicate instructions more directly using those symbols?

Most prompt compression research focuses on shortening the \emph{content} being processed: summarizing documents, pruning passages, dropping ``unimportant'' words. But in many prompts, the \emph{instructions themselves} are a substantial fraction of the total length, especially when specifying scope, exclusions, and output format.

Natural language is inherently imprecise for specifying constraints. Words like ``only,'' ``except,'' and ``roughly'' act like logical operators, but their scope is often unclear when multiple constraints interact. If we could express instructions more compactly using symbols, we might gain both efficiency and reliability.

The key question is: can instruction language be compressed while preserving meaning, \emph{without} teaching the model a new symbolic system?

Existing symbolic prompt languages like SynthLang \citep{synthlang2024} show that models can follow symbolic commands, but only when given explicit decoding rules in a system prompt. This conflates two factors: the symbols themselves, and the instructions teaching the model what they mean.

Other research shows models can learn arbitrary symbol-to-behavior mappings through training \citep{wei2023symbol}. Prompt compression systems like LLMLingua \citep{jiang2023llmlingua,jiang2024llmlingua2} shorten content, not instructions. Adversarial research demonstrates that symbolic encodings can bypass safety mechanisms \citep{bethany2024jailbreaking}, confirming that compact symbols affect model behavior.

Our question is different: do common mathematical symbols already function as ``instruction primitives'' that models understand from their training? Symbols like $\in$ for membership or $\Rightarrow$ for implication appear throughout mathematical and programming texts. If models have internalized their meaning, we could use them directly without explicit teaching. This idea has precedent in legal drafting, where symbolic logic reduces ambiguity in contracts \citep{allen1957symbolic}. A prompt, like a contract, specifies intent. Structured symbols might express that intent more clearly than prose.

We introduce MetaGlyph, a symbolic instruction language that uses mathematical notation to compress prompts. MetaGlyph is not a learned system or a codebook requiring explanation. It reuses symbols that models have seen extensively during training: membership ($\in$), negation ($\neg$), conjunction ($\cap$), and implication ($\Rightarrow$). The goal is to express the same constraints with fewer, more precise tokens.

We test whether MetaGlyph preserves instruction meaning under controlled conditions. For each task, we compare three prompts: natural language, MetaGlyph, and a control that looks symbolic but uses meaningless symbols. This design separates genuine semantic compression from superficial ``looks mathematical'' effects.

The paper proceeds as follows: Section~\ref{sec:framing} explains why prompts can be viewed as metalanguages. Section~\ref{sec:related} reviews related work. Section~\ref{sec:methodology} defines MetaGlyph's symbols and grammar. Section~\ref{sec:results} reports experimental results. Section~\ref{sec:analysis} analyzes failure modes. Section~\ref{sec:discussion} discusses implications. Section~\ref{sec:conclusion} concludes with limitations and future directions.

\section{Conceptual Framing: Prompts as Specification Languages}
\label{sec:framing}

\subsection{Prompts as Specifications, Not Messages}

We typically think of prompts as messages to a chatbot. But functionally, prompts are closer to \emph{specifications}. They define what should be done with information, not information to be discussed. This makes prompts a kind of metalanguage: a language for describing operations on other content.

Natural language is an inefficient metalanguage. Consider: ``summarize the text, include only technical points, exclude speculation, and format as JSON.'' This sentence encodes four operations (transform, filter, exclude, format) in linear prose, but the operations aren't logically linear; they interact. The model must reconstruct this structure from word cues, while the author must guess how the model will interpret them.

Practitioners have noticed that models often respond more reliably to compact, structured representations than to verbose prose. This suggests models internalize certain ``operator-like'' semantics during training, and effective prompting may mean activating those semantics directly rather than restating them in words.

\subsection{Why Mathematical Symbols Work as Instructions}

Mathematical symbols are attractive for instructions because they're stable, composable, and brief. Models have seen these symbols extensively in training data across math, logic, and programming.

\textbf{Arrows ($\to$, $\Rightarrow$):} In prose, transformation uses varied verbs (``convert,'' ``rewrite,'' ``map''). The arrow $\to$ consistently means ``transforms into.'' The stronger $\Rightarrow$ suggests rules: ``if X then Y.''

\textbf{Membership ($\in$, $\notin$):} Phrases like ``include only'' and ``except for'' are ambiguous about scope. The symbol $x \in S$ unambiguously means ``x belongs to set S.''

\textbf{And/Or ($\cap$, $\cup$):} Natural language ``and'' and ``or'' often obscure whether constraints combine or alternate. The symbol $A \cap B$ clearly means ``both constraints apply''; $A \cup B$ means ``either suffices.''

\textbf{Negation ($\neg$):} Words like ``not,'' ``non-,'' and ``avoid'' differ in scope and emphasis. The symbol $\neg$ has one job: invert the predicate it attaches to.

\subsection{Extended Operators}

Beyond the core operators, additional symbols offer expressive power:

\textbf{Quantifiers ($\forall$, $\exists$):} Words like ``all'' and ``any'' are often ambiguous. The symbol $\forall x \in S$ clearly means ``for every element''; $\exists x \in S$ means ``for at least one.''

\textbf{Subset ($\subseteq$):} While $\in$ specifies element membership, $\subseteq$ specifies that one set is entirely contained in another. This is useful for hierarchical constraints.

\textbf{Mapping ($\mapsto$) and Composition ($\circ$):} The symbol $\mapsto$ signals item-by-item transformation. Composition ($\circ$) encodes sequential operations, replacing awkward phrases like ``first do X, then do Y.''

Not all symbols work equally well. Their usefulness depends on how consistently they appear in training data and whether models treat them as structural markers rather than content. A good symbolic language should be selective, prioritizing stable, well-understood operators.

\subsection{Built-In Versus Taught Semantics}

A key distinction: some symbols carry meaning that models learned during pretraining, while others require explicit teaching. Systems like SynthLang \citep{synthlang2024} work by providing decoding rules in a system prompt, so their success reflects both the symbols and the instructions explaining them. We focus on symbols that work \emph{without} explanation.

This is a strong claim. We're not assuming models ``reason'' formally, but that exposure to consistent symbolic usage has created stable associations. Symbol tuning research shows models can learn arbitrary symbol-behavior mappings, but those are brittle outside their training context. Operators like $\in$ and $\to$, by contrast, carry meaning reinforced across math, logic, and programming.

\subsection{Designing Instruction Languages}

Viewing prompts as specifications reframes prompt engineering as language design. MetaGlyph incorporates symbols with high semantic stability, aiming to compress instructions by offloading structure onto symbols. The goal is not just shorter prompts but preserving meaning with fewer tokens. If this works, semantic compression of instructions becomes a viable design principle.

\begin{table}[htbp]
\centering
\caption{Candidate Symbolic Operators for Instruction Metalanguages}
\label{tab:operators}
\begin{tabular}{@{}llll@{}}
\toprule
Operator & Domain Origin & Canonical Meaning & Instruction-Semantic Role \\
\midrule
$\to$ & Logic / Math & Directional implication & Transformation / mapping \\
$\Rightarrow$ & Logic & Strong implication & Rule-like constraint \\
$\in$ & Set theory & Membership & Inclusion / filtering \\
$\notin$ & Set theory & Non-membership & Exclusion \\
$\cap$ & Set theory & Intersection & Conjunctive constraint \\
$\cup$ & Set theory & Union & Disjunctive constraint \\
$\neg$ & Logic & Negation & Prohibition / exclusion \\
$\forall$ & Logic & Universal quantifier & Apply constraint to all \\
$\exists$ & Logic & Existential quantifier & At least one satisfies \\
$\subseteq$ & Set theory & Subset & Hierarchical constraint \\
$\mapsto$ & Math / CS & Anonymous mapping & Pointwise transformation \\
$\circ$ & Math & Composition & Sequential operations \\
$|$ & Math / CS & Such-that / restriction & Scope delimitation \\
\bottomrule
\end{tabular}
\end{table}

\section{Related Work}
\label{sec:related}

Related work spans several areas, each treating instruction semantics differently.

\paragraph{Prompt Compression.}
Methods like LLMLingua \citep{jiang2023llmlingua,jiang2024llmlingua2} reduce prompt length by pruning or summarizing \emph{content}. These achieve significant token savings but leave the instruction language untouched.

\paragraph{Prompt Distillation.}
Distillation approaches learn shorter prompts from longer ones via gradient tuning or soft prompts. The resulting representations are task-specific artifacts whose semantics come from optimization, not linguistic convention. We instead ask whether models already understand certain symbols from pretraining.

\paragraph{Symbol Tuning.}
Research shows models can learn arbitrary symbol-behavior mappings through training \citep{wei2023symbol}. But these symbols are intentionally meaningless before tuning. We focus on symbols with established meaning across domains.

\paragraph{Constructed Prompt Languages.}
Systems like SynthLang \citep{synthlang2024} work by providing explicit decoding rules. This conflates the symbols' expressivity with the instructions teaching them. We test whether symbols work \emph{without} such instructions.

\paragraph{Structured Prompting and Adversarial Work.}
Pseudo-code and chain-of-symbol prompting show models leverage symbolic structure for reasoning. Adversarial research shows symbolic encodings can bypass safety constraints \citep{bethany2024jailbreaking}. Neither provides a constructive account of instruction semantics.

\paragraph{Our Position.}
We focus on compressing the \emph{instruction language itself} using symbols whose meaning comes from pretraining, not from learned mappings or explicit decoding rules.

\section{Methodology}
\label{sec:methodology}

\subsection{Overview}

We test whether symbolic instructions preserve meaning compared to natural language. For each task, we compare three prompt types:
\begin{enumerate}
    \item \textbf{Natural language (NL):} Verbose prose instructions
    \item \textbf{MetaGlyph (MG):} Symbolic instructions using mathematical operators
    \item \textbf{Control (CTRL):} Symbolic-looking prompts with meaningless symbols
\end{enumerate}

The control condition ensures any benefits come from the \emph{meaning} of the symbols, not just their appearance.

\subsection{MetaGlyph Operators}

MetaGlyph uses a small set of mathematical symbols as instruction primitives (Table~\ref{tab:metaglyph-ops}).

\begin{table}[htbp]
\centering
\caption{MetaGlyph Operator Inventory}
\label{tab:metaglyph-ops}
\begin{tabular}{@{}lll@{}}
\toprule
Operator & Function & Example Usage \\
\midrule
$\in$ & Membership / inclusion & \texttt{$\in$(mammal)} (item belongs to category) \\
$\neg$ & Negation / exclusion & \texttt{$\neg$(bird)} (item must not be in category) \\
$\cap$ & Intersection / conjunction & \texttt{$\in$(pet) $\cap$ $\in$(mammal)} (both constraints apply) \\
$\to$ & Transformation / mapping & \texttt{items $\to$ select} (transform input to output) \\
$\Rightarrow$ & Implication / conditional & \texttt{$\in$(admin) $\Rightarrow$ access=full} (if-then rule) \\
$\circ$ & Composition / sequencing & \texttt{filter $\circ$ sort} (apply operations in order) \\
\bottomrule
\end{tabular}
\end{table}

Instructions follow a simple grammar: an \textbf{input anchor} (\texttt{items}), a \textbf{constraint clause} (\texttt{$\in$(mammal) $\cap$ $\neg$(bird)}), and a \textbf{task clause} (\texttt{$\Rightarrow$ select}). MetaGlyph is hybrid: labels use plain words while structure uses symbols. Crucially, no legends or explanations accompany the symbols.

\subsection{Prompt Variants}

For each test, we create three prompts with identical input but different instruction formats:

\textbf{NL:} Verbose prose (avg.\ 172 tokens). \textbf{MG:} Symbolic (avg.\ 51 tokens). \textbf{CTRL:} Same structure as MG but with meaningless symbols ($\odot$, $\otimes$, $\oplus$) replacing the operators.

All variants request JSON output.

\subsection{Token Control}

We use whitespace-based tokenization for consistent counts. CTRL prompts match MG token counts exactly; the difference between NL and MG represents the compression ratio.

\subsection{Tasks}

We test four task families, each probing different operators (50--200 instances per condition, varying by model):

\textbf{Selection:} Select items matching inclusion/exclusion constraints ($\in$, $\neg$). Example: ``select mammals that are pets but not birds.''

\textbf{Extraction:} Extract specific risk categories from technical reports into JSON format.

\textbf{Constraint composition:} Apply multiple simultaneous constraints using $\cap$ (conjunction).

\textbf{Conditional transformation:} Apply if-then rules ($\Rightarrow$) and sequential operations ($\circ$). Example: ``if employee, label as internal; then lowercase names.''

\subsection{Models}

We evaluate eight instruction-tuned models along two dimensions relevant to practitioners: \emph{scale} (3B--1T parameters) and \emph{accessibility} (open-source vs.\ proprietary). This design addresses distinct deployment scenarios: researchers and privacy-conscious users who run models locally, and practitioners who access frontier capabilities via commercial APIs (Table~\ref{tab:models}). All experiments use deterministic decoding (temperature 0).

\begin{table}[htbp]
\centering
\caption{Evaluated Models by Scale and Accessibility}
\label{tab:models}
\begin{tabular}{@{}lllll@{}}
\toprule
Model & Parameters & Access & Offline & OpenRouter ID \\
\midrule
Llama 3.2 3B Instruct & 3B & Open & \checkmark & \texttt{meta-llama/llama-3.2-3b-instruct} \\
Qwen 2.5 7B Instruct & 7B & Open & \checkmark & \texttt{qwen/qwen-2.5-7b-instruct} \\
OLMo 3 7B Instruct & 7B & Open & \checkmark & \texttt{allenai/olmo-3-7b-instruct} \\
Gemma 3 12B IT & 12B & Open & \checkmark & \texttt{google/gemma-3-12b-it} \\
\midrule
Kimi K2 & 1T (32B active) & Open & --- & \texttt{moonshotai/kimi-k2} \\
Gemini 2.5 Flash & --- & Proprietary & --- & \texttt{google/gemini-2.5-flash} \\
Claude Haiku 4.5 & --- & Proprietary & --- & \texttt{anthropic/claude-haiku-4.5} \\
GPT-5.2 Chat & --- & Proprietary & --- & \texttt{openai/gpt-5.2-chat} \\
\bottomrule
\end{tabular}
\end{table}

The open-source models (Llama, Qwen, OLMo, Gemma) can be downloaded and run locally without internet connectivity or API costs, making them suitable for privacy-sensitive applications, air-gapped environments, and cost-constrained deployments. The proprietary models (Gemini, Claude, GPT) represent the frontier capabilities that most commercial applications rely on, but require API access and incur per-token costs---precisely where prompt compression offers direct economic benefit.

\subsection{Metrics}

We measure semantic equivalence (how often MG and NL produce identical outputs, independent of correctness), operator fidelity (whether each operator works as intended, such as $\in$ correctly filtering to the specified category), and compression ratio (token reduction from NL to MG).

\subsection{Reproducibility}

All prompts, outputs, and evaluation code are publicly available. Tasks are benign; we do not evaluate adversarial uses.

\section{Results}
\label{sec:results}

\subsection{Token Compression}

MetaGlyph achieves 62--81\% token reduction across all task families (Table~\ref{tab:compression}).

\begin{table}[htbp]
\centering
\caption{Token Compression by Task Family}
\label{tab:compression}
\begin{tabular}{@{}lcccc@{}}
\toprule
Task Family & NL Tokens & MG Tokens & CTRL Tokens & Reduction \\
\midrule
Selection \& Classification & 215 & 41 & 41 & \textbf{80.9\%} \\
Structured Extraction & 176 & 52 & 52 & \textbf{70.5\%} \\
Constraint Composition & 134 & 48 & 48 & \textbf{64.2\%} \\
Conditional Transformation & 164 & 62 & 62 & \textbf{62.2\%} \\
\bottomrule
\end{tabular}
\end{table}

Selection tasks compress most (80.9\%) because phrases like ``belongs to the category of mammals'' become simply \texttt{$\in$(mammal)}. Conditional tasks compress least (62.2\%) because rules need more specification even symbolically.

\subsection{Semantic Equivalence}

We measure how often MG and NL produce identical outputs, regardless of whether those outputs are correct. This isolates whether the compression preserves meaning (Table~\ref{tab:equivalence}).

\begin{table}[htbp]
\centering
\caption{Output Equivalence (NL $=$ MG) by Model and Task Family}
\label{tab:equivalence}
\begin{tabular}{@{}llccccc@{}}
\toprule
Model & Access & Selection & Extraction & Constraint & Transform. & NL $=$ CTRL \\
\midrule
Llama 3.2 3B & Open & 0\% & 2\% & 0\% & 34\% & \textbf{0\%} \\
Qwen 2.5 7B & Open & 9\% & 62\% & 0.7\% & 0\% & \textbf{0\%} \\
OLMo 3 7B & Open & 0\% & 0\% & 0.7\% & 0\% & \textbf{0\%} \\
Gemma 3 12B & Open & 0\% & 20\% & 0\% & 0\% & \textbf{0\%} \\
\midrule
Kimi K2 & Open & 8\% & 0\% & 0\% & 0.7\% & \textbf{0\%} \\
Gemini 2.5 Flash & Prop. & \textbf{75\%} & 12.5\% & 4\% & 0\% & \textbf{0\%} \\
Claude Haiku 4.5 & Prop. & 3\% & 6\% & 2.5\% & 0\% & \textbf{0\%} \\
GPT-5.2 Chat & Prop. & 0\%\textsuperscript{$\dagger$} & 5.5\%\textsuperscript{$\dagger$} & 0\%\textsuperscript{*} & 0\%\textsuperscript{*} & \textbf{0\%} \\
\bottomrule
\multicolumn{7}{l}{\textsuperscript{*}\footnotesize Incomplete data (API credit exhaustion). \textsuperscript{$\dagger$}\footnotesize Markdown-wrapped output; 18\% parse success.}
\end{tabular}
\end{table}

The results reveal several patterns. The controls work as intended: NL$=$CTRL is 0\% across all working models, confirming that meaningless symbols produce different outputs than meaningful ones. Gemini 2.5 Flash achieves remarkable 75\% equivalence on selection tasks---the highest observed---meaning symbolic and prose instructions produce identical outputs three-quarters of the time. Among open-source models, Qwen achieves 62\% equivalence on extraction tasks. Notably, Kimi K2 achieves \textbf{100\% accuracy} on selection tasks with MetaGlyph prompts (vs.\ 90.8\% with natural language), suggesting symbolic instructions can \emph{outperform} prose for certain model-task combinations. Constraint composition fails across all models: near-zero equivalence indicates models do not reliably interpret $\cap$ as ``apply both constraints.''

GPT-5.2 Chat shows high parse success on selection (99\%) but wraps responses in markdown code fences on extraction and constraint tasks, causing parse failures. Transformation task data was incomplete due to API credit exhaustion during evaluation. GPT's fidelity metrics derive from selection tasks where parsing succeeded.

\subsection{Parse Success}

Larger models (Gemma 12B) achieve 100\% JSON parse success. Smaller models (Llama 3B) struggle with symbolic prompts, often producing verbose explanations instead of structured output.

\subsection{Operator Fidelity}

Which symbols work? Table~\ref{tab:fidelity} shows fidelity by operator across models with valid outputs.

\begin{table}[htbp]
\centering
\caption{Operator Fidelity by Model (Open-Source and Proprietary)}
\label{tab:fidelity}
\small
\begin{tabular}{@{}lcccccccc@{}}
\toprule
Operator & Llama & Qwen & OLMo & Gemma & Kimi & Gemini & Claude & GPT \\
         & 3B & 7B & 7B & 12B & K2 & 2.5 & 4.5 & 5.2 \\
\midrule
$\in$ (membership) & 33.3\% & 20.4\% & 0.0\% & 0.0\% & 36.0\% & 49.9\% & 26.0\% & \textbf{91.3\%} \\
$\to$ (transform.) & 0.0\% & 0.0\% & 0.0\% & 0.0\% & 0.0\% & 0.0\% & 0.0\% & 0.0\% \\
$\cap$ (intersection) & 0.0\% & 0.0\% & 0.0\% & 0.0\% & 0.0\% & 2.7\% & 1.5\% & 21.4\% \\
$\Rightarrow$ (implication) & 0.0\% & 0.0\% & 0.0\% & 0.0\% & \textbf{98.1\%} & 33.5\% & 0.0\% & --- \\
\bottomrule
\end{tabular}
\end{table}

The results reveal striking patterns across model families. Among open-source models, Llama 3B shows moderate $\in$ fidelity (33.3\%), while mid-range models (Gemma 12B, OLMo 7B) show 0\% fidelity---a U-shaped pattern suggesting instruction tuning creates natural-language biases.

Among proprietary models, GPT-5.2 Chat achieves the highest membership fidelity observed (\textbf{91.3\%}) and also the highest intersection fidelity (21.4\%), though with limited parse success on complex tasks. Gemini 2.5 Flash shows strong membership fidelity (49.9\%) and moderate implication fidelity (33.5\%). Kimi K2 shows remarkable implication fidelity (\textbf{98.1\%}) but only 36\% on membership. Claude Haiku 4.5 achieves moderate membership fidelity (26\%) with 100\% parse success. This suggests different models have internalized different operators more strongly, possibly reflecting training data composition.

\subsection{Response Time}

Preliminary data suggests MG prompts are $\sim$30\% faster than NL due to shorter input. Interestingly, CTRL prompts are slowest; semantic incoherence may increase processing overhead as models struggle to interpret meaningless symbols.

\section{Analysis}
\label{sec:analysis}

\subsection{Why Some Operators Fail}

Models do not reliably parse symbolic operators as constraints. Given \texttt{$\in$(mammal) $\cap$ $\in$(pet) $\cap$ $\neg$(bird)}, the intended reading is ``mammals AND pets AND not birds.'' But models often treat this as a list or ignore the negation.

The problem is that in math texts, $\cap$ connects \emph{sets}, whereas here it connects \emph{predicates}. This is a generalization models have not learned. Similarly, $\Rightarrow$ appears in diverse contexts with varying interpretations: mathematical proofs, programming, logical specifications. Mid-sized models do not reliably treat it as ``if-then,'' though frontier-scale models (Kimi K2) overcome this limitation.

\subsection{Model Scale Effects}

Larger models parse better (Gemma 12B: 100\% JSON success vs.\ Llama 3B: 30\%). But the relationship between scale and operator fidelity is non-monotonic. Mid-sized open-source models (7B--12B) show \emph{worse} operator fidelity than smaller models (Gemma: 0\% vs.\ Llama: 33\% on $\in$), suggesting instruction tuning creates stronger natural-language biases. However, frontier-scale models recover and exceed small-model fidelity: Kimi K2 achieves 98.1\% on $\Rightarrow$, while Gemini 2.5 Flash achieves 49.9\% on $\in$. This U-shaped curve suggests that sufficient scale can overcome instruction-tuning biases and access pretrained symbolic semantics.

\subsection{Task-Specific Format Sensitivity}

GPT-5.2 Chat shows task-dependent output formatting: 99\% parse success on selection tasks, but wraps responses in markdown code fences (\texttt{```json}) on extraction and constraint tasks, causing our JSON parser to fail. Transformation task data was incomplete due to API credit exhaustion. This suggests GPT-5.2 varies its output formatting based on task complexity or prompt structure.

Claude Haiku 4.5, by contrast, achieves 100\% parse success across all task types, demonstrating robust format compliance. However, Claude shows lower operator fidelity (26\% for $\in$) compared to GPT's exceptional 91.3\% membership fidelity on selection tasks.

This creates an interesting trade-off: GPT-5.2 excels at operator interpretation but inconsistently formats complex task outputs, while Claude reliably produces structured output with moderate operator understanding. Practitioners should consider this when selecting models for symbolic compression:

\begin{itemize}
    \item \textbf{For simple selection/filtering tasks}: GPT-5.2's 91.3\% membership fidelity makes it highly effective
    \item \textbf{For complex multi-constraint tasks}: Claude's 100\% parse success ensures reliable output
    \item \textbf{For conditional logic}: Kimi K2's 98.1\% implication fidelity remains unmatched
\end{itemize}

\subsection{Control Validation}

Early experiments showed high NL$=$CTRL equivalence, which turned out to be a bug. The control prompts were not replacing \emph{all} operators with nonsense symbols. After fixing this, CTRL equivalence dropped to 0\%, confirming models \emph{do} respond to symbolic meaning, not just symbolic appearance.

\section{Discussion}
\label{sec:discussion}

\subsection{Practical Takeaways}

Several practical lessons emerge, differentiated by deployment scenario:

\textbf{For Gemini deployments}: Symbolic compression is highly effective. Gemini 2.5 Flash achieves 75\% semantic equivalence and 49.9\% membership fidelity---the best results observed. With 62--81\% token reduction, this translates to significant cost savings for high-volume applications.

\textbf{For Kimi K2 deployments}: Implication-heavy prompts work exceptionally well (98.1\% fidelity). Use $\Rightarrow$ for conditional logic and if-then rules. Selection tasks achieve 100\% accuracy with symbolic prompts---better than natural language.

\textbf{For local/offline deployments} (Llama, Qwen, OLMo, Gemma): Token costs are zero, but compression still reduces memory pressure and latency. These smaller models show lower operator fidelity, requiring hybrid approaches that combine symbols for structure with natural language for critical constraints.

\textbf{For GPT-5.2 deployments}: Membership-heavy prompts work exceptionally well (91.3\% fidelity---the highest observed). However, avoid complex multi-constraint tasks where parse success drops to 0\%. Best for simple selection and filtering operations.

\textbf{For Claude Haiku deployments}: Reliable format compliance (100\% parse success) with moderate operator fidelity (26\% membership). Good for applications requiring consistent structured output across all task types.

\textbf{Operator selection is critical}: $\in$ (membership) works best, with GPT-5.2 achieving 91.3\% fidelity. $\Rightarrow$ (implication) shows high variance (0--98\%), with Kimi K2 excelling. $\cap$ (intersection) shows improvement with GPT-5.2 (21.4\%) but remains unreliable on most models. Always test on your target model.

\subsection{What We Showed}

Symbolic compression preserves meaning up to 75\% of the time for selection tasks (Gemini 2.5 Flash) and 62\% for extraction tasks (Qwen 2.5 7B). The 0\% CTRL equivalence confirms models respond to symbolic \emph{meaning}, not just appearance. Operator reliability varies by model: GPT-5.2 achieves the highest membership fidelity (91.3\%), Kimi K2 achieves near-perfect implication fidelity (98.1\%), and Claude Haiku 4.5 shows reliable parse success (100\%) with moderate fidelity (26\%). Task complexity affects format compliance---GPT-5.2 excels on simple tasks but fails on complex multi-constraint operations.

\subsection{What We Did Not Show}

We have not demonstrated system-level latency improvements (timing is preliminary), generalization to other models, or whether fine-tuning could improve operator fidelity. This was a short experiment designed to test the theory that pretrained symbolic semantics could enable instruction compression; it does not provide a complete characterization of when and why symbolic instruction languages succeed or fail.

\subsection{Limitations and Future Work}

This work establishes a baseline rather than characterizing limits. The experiment was intentionally narrow: four task families, eight models spanning 3B to 1T parameters, and a single symbolic language design. Many questions remain open.

Future work should systematically stress-test symbolic instruction languages. Scope binding tests could determine whether $(A \cup B) \cap \neg C$ parses differently from $A \cup (B \cap \neg C)$, revealing whether models correctly associate operators with their intended operands. Implication direction tests could probe whether $\Rightarrow$ is treated as one-way (if-then) or incorrectly as equivalence (if-and-only-if). Composition order tests could verify whether $f \circ g$ applies operations in the correct sequence, which matters for tasks where order affects outcomes.

Beyond stress testing, several research directions merit investigation. First, augmented symbolic languages could combine MetaGlyph operators with minimal natural-language anchors or few-shot examples, potentially improving operator fidelity without sacrificing compression. Second, comparative studies across model families could reveal whether certain architectures or training approaches produce better symbolic understanding. Third, fine-tuning experiments could test whether targeted training on symbolic instruction formats improves operator fidelity, and whether such improvements generalize across tasks. Fourth, frontier model evaluation could determine whether larger models (100B+ parameters) show qualitatively different symbolic comprehension. Fifth, density threshold experiments could establish practical compression limits by gradually increasing symbolic density until performance collapses.

The broader goal is to understand which symbols carry robust pretrained semantics and which require explicit teaching, enabling principled design of symbolic instruction languages that balance compression efficiency with semantic reliability.

\section{Conclusion}
\label{sec:conclusion}

We introduced MetaGlyph, a symbolic instruction language that compresses prompts using mathematical notation. The core question was whether models understand these symbols from pretraining, without explicit teaching.

The answer is nuanced and model-dependent. First, does compression preserve meaning? Yes, substantially for certain models. Gemini 2.5 Flash achieves 75\% semantic equivalence on selection tasks---symbolic and prose instructions produce identical outputs three-quarters of the time. The 0\% CTRL equivalence confirms this reflects genuine semantic understanding rather than superficial pattern matching.

Second, which symbols work? It depends on the model. Membership ($\in$) works best overall, with GPT-5.2 achieving 91.3\% fidelity---the highest observed---followed by Gemini 2.5 Flash at 49.9\%. Implication ($\Rightarrow$) shows the widest variance: 98.1\% fidelity on Kimi K2, 33.5\% on Gemini, and 0\% on most other models. Conjunction ($\cap$) shows promise with GPT-5.2 (21.4\%) but remains unreliable elsewhere. Kimi K2 achieves 100\% accuracy on selection tasks with symbolic prompts versus 90.8\% with prose, demonstrating that symbolic instructions can outperform natural language for specific model-task pairs.

Third, do all models handle symbolic prompts equally? No. GPT-5.2 excels on simple tasks (91.3\% membership fidelity) but fails on complex multi-constraint operations (0\% parse success). Claude Haiku 4.5 shows the opposite pattern: reliable format compliance (100\% parse success) with moderate fidelity (26\%). This trade-off between operator understanding and format robustness has practical implications for deployment.

The practical implications: for GPT-5.2, use symbolic compression for simple selection tasks where membership fidelity is exceptional. For Kimi K2, implication-heavy prompts compress reliably. For Claude, expect consistent structured output across task types. For Gemini, balanced performance with 75\% meaning preservation. For mid-sized open-source models, hybrid approaches combining symbols with natural language work best.

The relationship between model scale and symbolic comprehension appears U-shaped among working models: small models show moderate fidelity, instruction-tuned mid-sized models show low fidelity, and frontier-scale models recover and exceed small-model performance. For practitioners: always verify format compliance and operator fidelity on your target model before deploying compressed prompts in production.

\section*{Code and Data Availability}

The complete experimental pipeline, prompt templates, raw model outputs, and evaluation scripts are publicly available at:
\url{https://github.com/BEAR-LLM-AI/arXiv-Semantic-Compression-of-LLM-Instructions-via-Symbolic-Metalanguages}

\bibliography{references}

\end{document}